\documentclass{article}

    \usepackage[final]{mlrsa_2020}

\usepackage[utf8]{inputenc} 
\usepackage[T1]{fontenc}    
\usepackage{hyperref}       
\usepackage{url}            
\usepackage{booktabs}       
\usepackage{amsfonts}       
\usepackage{nicefrac}       
\usepackage{microtype}      

\usepackage{times}
\usepackage{latexsym}
\usepackage{amsmath}
\usepackage{amssymb}
\usepackage{color}
\usepackage{graphicx}
\usepackage{mathtools}

\usepackage{subcaption}

\usepackage{titlesec}

\newcommand{\repeatthanks}{\textsuperscript{\thefootnote}}

\title{Understanding Attention: In Minds and Machines}

\author{%
Shriraj P. Sawant\thanks{equal contribution} \\
Dept. of Computer Science and Engineering\\
Indian Institute of Technology Gandhinagar\\
\texttt{sawant\_shriraj@iitgn.ac.in} \\
\And 
Shruti Singh\repeatthanks \\
Dept. of Computer Science and Engineering\\
Indian Institute of Technology Gandhinagar\\
\texttt{singh\_shruti@iitgn.ac.in} \\
}

\begin{document}
\maketitle
\begin{abstract}
Attention is a complex and broad concept, studied across multiple disciplines spanning artificial intelligence, cognitive science, psychology, neuroscience, and related fields. Although many of the ideas regarding attention do not significantly overlap among these fields, there is a common theme of adaptive control of limited resources. In this work, we review the concept and variants of attention in artificial neural networks (ANNs). We also discuss the origin of attention from the neuroscience point of view parallel to that of ANNs. Instead of having seemingly disconnected dialogues between varied disciplines, we suggest grounding the ideas on common conceptual frameworks for a systematic analysis of attention and towards possible unification of ideas in AI and Neuroscience.
\end{abstract}

\section{Introduction}
Attention mechanism in artificial neural networks was first proposed for the task of Machine Translation by \citet{DBLP:journals/corr/BahdanauCB14} and is now widely applied to various other tasks in natural language processing and computer vision such as question answering, summarization, document classification, image captioning, image classification, and others. Simultaneously, attention in biology has been widely studied to understand the flexibility of controlling limited computational resources in the brain. In this paper, we propose to study the attention mechanism in neural networks and the human brain. 

The basic attention mechanism proposed by \citet{DBLP:journals/corr/BahdanauCB14} for machine translation works by learning to align the words in the target language with words in the source language and by learning a language model for the target language. The target sequence is generated based on the previous word and a context which is fundamentally a mapping between the words in the source and the target language. \citet{garcia2013brain} show that language interpreters interpret the sentence in one language and then translate it to the other language, rather than translating on a word-by-word basis. The language acquisition process in bilinguals studied by \citet{ullman2015declarative}, puts forward a similar Declarative/Procedural Model in which different types of memories are responsible for learning the syntax of the two languages and the semantic knowledge about the words and concepts.

The rest of the paper is organized as follows. In Section~\ref{sec:Att-NN}, we review the attention models in neural networks. The language acquisition process in bilinguals is presented in Section~\ref{sec:TransInBrain}. Section~\ref{sec:Att-Neuro} describes the attention mechanism in humans from the neuroscience perspective. In Section~\ref{sec:UnifyingAINS}, we try to ground the parallels between artificial intelligence and neuroscience views of attention using the conceptual framework of Marr's levels of analysis.

\section{Attention in Artificial Neural Networks}
\label{sec:Att-NN}
\subsection{Traditional Encoder-Decoder for Language Translation}
\begin{figure}[!h]
    \centering
    \begin{minipage}{0.45\textwidth}
        \centering
        \includegraphics[width=0.5\textwidth, height=4.4cm]{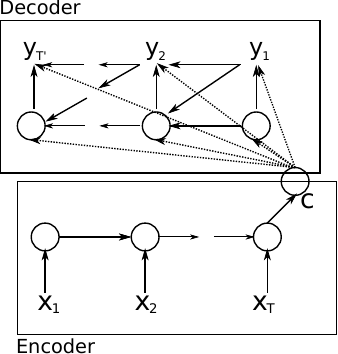}
        \caption{Basic Encoder-Decoder network with a single summary vector, image taken from \citet{Cho2014LearningPR}}
        \label{fig:basicencdec}
    \end{minipage}\hfill
    \begin{minipage}{0.45\textwidth}
        \centering
        \includegraphics[width=0.5\textwidth, height=4cm]{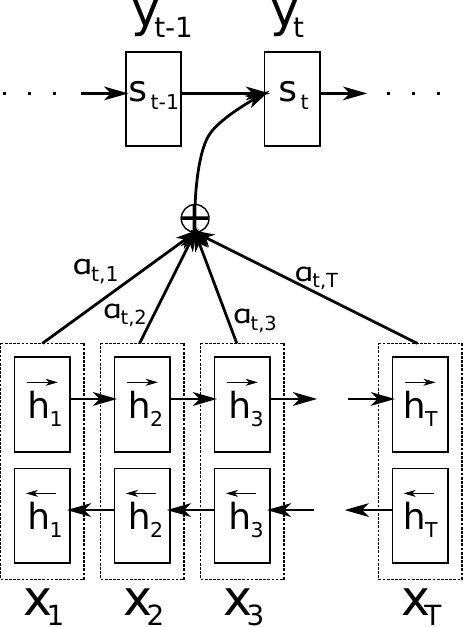}
        \caption{Attention architecture, image taken from \citet{DBLP:journals/corr/BahdanauCB14}}
        \label{fig:attentionarchimg}
    \end{minipage}
\end{figure}

A basic encoder-decoder framework was proposed by \citet{Cho2014LearningPR} for translation. The idea is to use an encoder network to read the input sequence and obtain a fixed representation of the sentence. Then the decoder network is used to get the output sequence from the previously obtained vector. A summary vector is generated from the hidden state of the encoder RNN after reading the whole sentence (detected by end-of-sequence marker) and is considered to contain the important information content of the input sentence. The structure of the network proposed by \citet{Cho2014LearningPR} is as follows:
\begin{align}
    \label{eq:trad-hidden}
    &h_t = f(h_{t-1}, y_{t-1}, c),
    \\
    \label{eq:trad-pred}
    &p(y_t | y_{t-1}, y_{t-2}, \dots, y{1}, c) = g(h_{t}, y_{t-1}, c)
\end{align}
where \emph{c} is the summary vector and \emph{c} = $h_{t}$. The output symbol at time \emph{t} is predicted from the hidden state at time \emph{t} i.e. $h_{t}$, previously generated output symbol $y_{t-1}$, and the summary \emph{c}. The network structure is presented in Figure~\ref{fig:basicencdec}. However, the performance of basic encoder-decoder approaches degrades as the length of the sentence increases.

\subsection{Basic Attention Mechanism}
The attention mechanism was proposed by \citet{DBLP:journals/corr/BahdanauCB14}. It works by trying to align words in the target and the source language and then translating. In \citet{DBLP:journals/corr/BahdanauCB14}, the encoder is a bi-directional RNN, and the decoder is an RNN. The probability of generating $t^{th}$ output word given the previously generated words and the input sequence x, is dependent on the word generated in the previous step, the hidden state of the decoder at time \emph{t} i.e. $s_{t}$, and the context vector $c_t$. $c_t$ is generated separately for each target word. Instead of having a single summary for all the output units as proposed in~\citet{Cho2014LearningPR}, a distinct context for each of the outputs is computed dynamically. 
\begin{equation}
    \label{eq:att-pred}
    p(y_t | y_{t-1}, y_{t-2}, \dots, y_{1}, c) = g(y_{t-1}, s_{t}, c_{t})
\end{equation}

The alignment model computes scores \emph{$e_{ij}$} which models how well the inputs at position j match with the output at position i.
\begin{align}
    \label{eq:att-eij}
    &e_{ij} = a(s_{i-1}, h_{j})
    \\
    \label{eq:att-alphaij}
    &\alpha_{ij} = \frac{exp(e_{ij})}{\sum_{k=1}^{T_x} exp(e_{ik})} 
\end{align}

\emph{a} is a simple feed forward network with a single hidden layer and $h_j$ is the hidden state of the bi-directional encoder RNN formed by the concatenation of the forward and the backward hidden states at step j. As it can be observed from equation~\ref{eq:att-alphaij}, the alignment scores are normalized. $\alpha_{ij}$ is the normalized alignment score of the $j^{th}$ input token with the decoder hidden state at $(i-1)$, with respect to all other tokens in the input sequence. It can also be interpreted as the probability that the $i^{th}$ word is translated from the source word $x_j$ during translation of the input sentence.

The context $c_i$, is the weighted sum of all the input hidden states for the $i^{th}$ output token. In fact, this is the expected hidden state as the $\alpha_{ij}$'s from equation~\ref{eq:att-alphaij} are probability values. In the proposed model, the next word is dependent on the word generated in the previous step, and the words that hidden state believes are relevant for generating the next word.  The architecture is presented in Figure~\ref{fig:attentionarchimg}.

The Alignment model proposed in \citet{DBLP:journals/corr/BahdanauCB14} is a feed forward MLP. Other variants of the alignment model are listed in Table~\ref{tab:alignmentModels}.

\begin{table}[h]
\centering
\begin{tabular}{|c|c|c|}
\hline

\textbf{Alignment Type}     & \textbf{Alignment} \\ \hline
Content            & $e_{ij}$ = cosine($s_{i-1}, h_j$) \\ \hline
Location           & $e_{ij}$ = softmax(W, $s_{i-1}$) \\ \hline
Dot Product        & $e_{ij}$ = $s_{i-1}^{T} . h_j$ \\ \hline
Scaled Dot Product & $e_{ij}$ = $\frac{s_{i-1}^{T} . h_j}{\sqrt(n)}$ \\ \hline

\end{tabular}
\caption{Various alignment model schemes}
\label{tab:alignmentModels}
\end{table}
\subsection{Attention Variants}
\begin{figure}
\centering
\begin{subfigure}{.5\textwidth}
    \centering
    \includegraphics[width=0.6\textwidth, height=4cm]{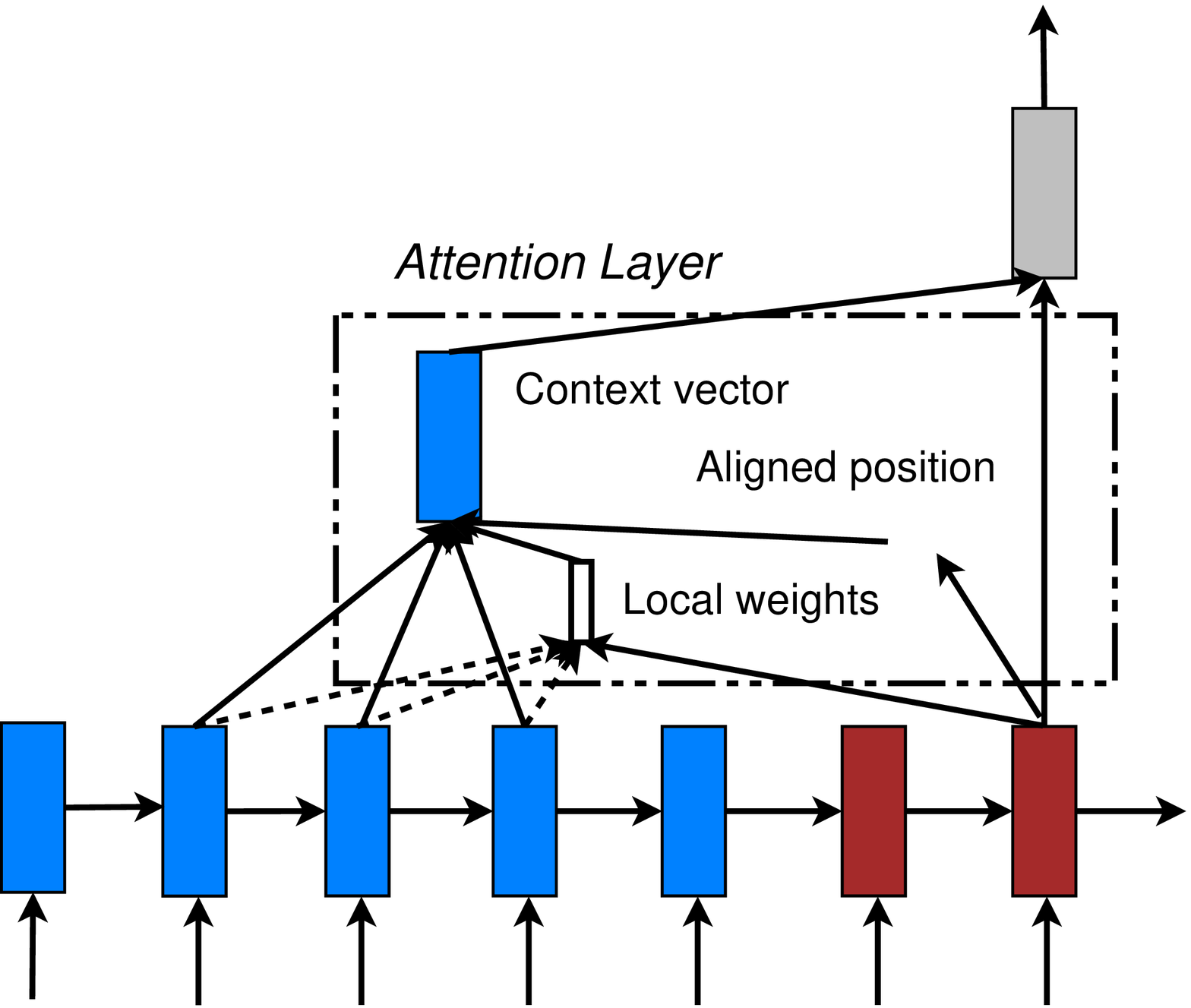}
    \caption{Local Attention}
    \label{fig:localatt}
\end{subfigure}%
\begin{subfigure}{.5\textwidth}
    \centering
    \includegraphics[width=0.6\textwidth, height=4cm]{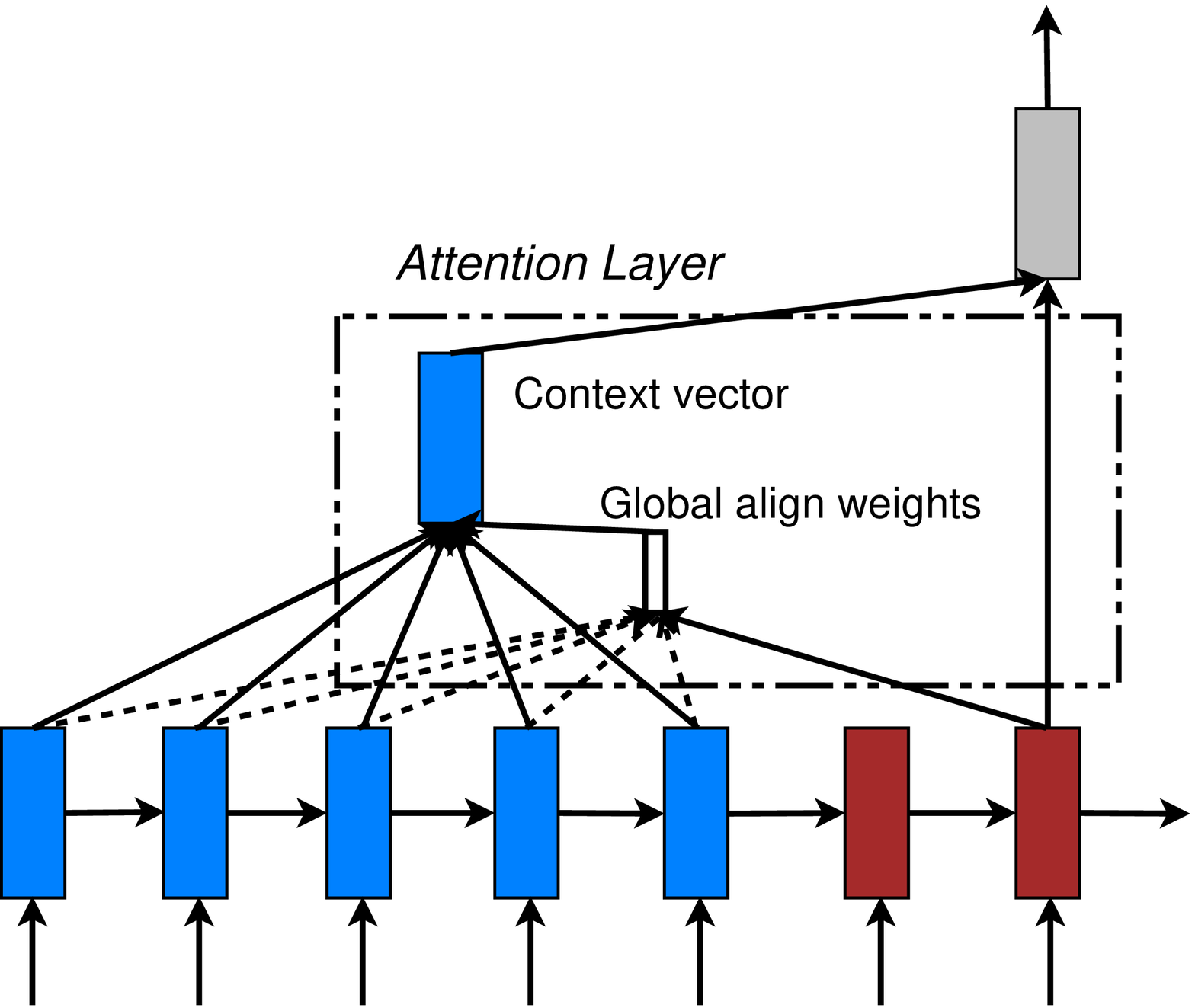}
    \caption{Global Attention}
    \label{fig:globalatt}
\end{subfigure}
\caption{Local and Global Attention, image taken from \citet{luong-etal-2015-effective}}
\label{fig:localglobalatt}
\end{figure}

\begin{figure}[h]
    \centering
    \includegraphics[width=0.5\textwidth]{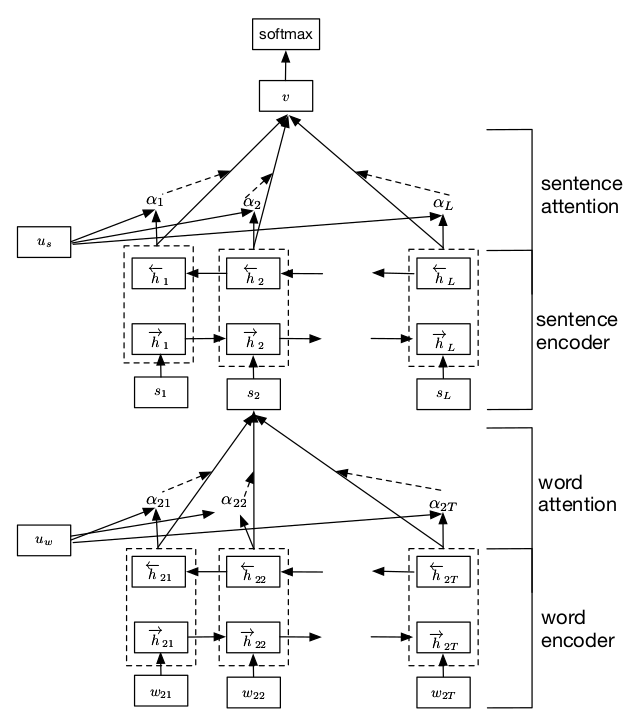}
    \caption{Hierarchical Attention, image taken from \citet{yang-etal-2016-hierarchical}}
    \label{fig:han_network}
\end{figure}

We discuss the variants of attention in this subsection. Broadly, categories of attention are:
\begin{enumerate}
    \item \textbf{Soft vs Hard Attention:}
    This was first proposed by \citet{xu2015show} for the task of generating image captions. The encoder is a convolution neural network, used to extract feature vectors from the image. The decoder is a long short-term memory network, used to caption the image by generating one word at a time. In Soft Attention, the alignment weights are placed all over the source image. On the other hand, Hard Attention selects one patch of the image to attend to at a time.
    \item \textbf{Local vs Global Attention:}
    \citet{luong-etal-2015-effective} proposed Local and Global Attention for the task of Machine Translation. Global Attention is similar to Soft Attention, and it considers all hidden states of the encoder when deriving the context. The context $c_t$ used for generating the target word at time step \emph{t}, is computed as the weighted average over all the source hidden states. Local Attention, instead of calculating a weighted average over the all the words, it selectively focus on a small context window. An aligned position $p_t$ is generated for each target word at time \emph{t}. The context $c_t$ is weighted average over [$p_t$ - D, $p_t$ + D]. The authors propose two types of alignment models: 
    \begin{itemize}
        \item Monotonic Alignment: It assumes that the source and the target sequences are monotonically aligned roughly. In this, the alignment position $p_t = t$ at time step \emph{t}.
        \item Predictive Alignment: The aligned position is learnt using a feed forward network. To favour points near $p_t$, the authors place a gaussian centered at $p_t$ with $\sigma = \frac{D}{2}$, during calculation of the alignment score.
    \end{itemize}
    The architecture taken from \citet{luong-etal-2015-effective} is presented in Figure~\ref{fig:localglobalatt}.
    \item \textbf{Self Attention: }
    Self Attention is the mechanism to capture different relations between words at different positions in the same sequence. Multi-head attention computes the attention multiple times parallely, which helps a model to learn information from multiple representation subspaces. Multi-head Self Attention is an integral component of transformer based models, which have been shown to perform very well on various natural language processing tasks. 
    \item \textbf{Hierarchical Attention: }
    Hierarchical Attention was proposed by \citet{yang-etal-2016-hierarchical} to take into account the hierarchical nature of the data. \citet{yang-etal-2016-hierarchical} propose a two-level attention mechanism - word-level attention in constructing sentence representations, and sentence-level attention in constructing document representation. It has been shown to be useful in document classification, summarization, and information extraction. The Hierarchical Attention network architecture is presented in Figure~\ref{fig:han_network}.
\end{enumerate}

\section{Language Translation in Human Brain}
\label{sec:TransInBrain}
In a study conducted on early bilinguals, \citet{garcia2013brain} show that word translations differ from sentence translations in human brain. Sentence translation involves syntactic processing and greater semantic and conceptual analysis demands. 
Previous studies of the brains of the human language translators have shown that humans first try to interpret the sentence in the source language and then translate it to the target language instead of translating on a word-by-word basis.

The Declarative/Procedural (DP) model is proposed in \citet{ullman2015declarative}. The Declarative Memory in humans is the memory that stores knowledge and facts. It consists of the Semantic and the Episodic memory. Semantic memory is responsible for the knowledge of words and concepts. Procedural memory is involved with learning how to perform different actions and skills such as riding a bike, tying shoe laces etc. 

The DP Model states that the Declarative Memory is responsible for storing distinctive knowledge about both languages in bilinguals. The knowledge of words, their phonological forms, hierarchical frames, inflectional forms, knowledge of proverbs and idioms of both the languages is stored in the Declarative Memory. The Procedural Memory is responsible for learning the syntax, especially for the first language. The learning abilities in procedural memory seem to be established early and then  decline, while declarative memory shows the opposite pattern. However, the second languages’ grammar is more dependent on the declarative memory as it learns faster. Since acquisition also depends on the age of the learner, it is likely that the syntax for the second language is acquired by the Declarative memory due to the decline in the learning abilities of the Procedural memory. Also, since the second language is exposed at an age later than the first language, the brain is already familiar with what syntax means.

It can be considered that the context vector proposed in \citet{DBLP:journals/corr/BahdanauCB14} learns the words and concepts in the language and a mapping of words in the two languages. This is similar to the knowledge stored in the Declarative Memory. The sequence-to-sequence architecture is responsible for learning the syntax, which works similar to the Procedural Memory.

\section{Attention in Neuroscience}
\label{sec:Att-Neuro}
\subsection{Overview of Attention}
As the rise of experimental psychology was inevitable, William James, the father of American psychology, famously declared that “Everyone knows what attention is” but unintentionally, such informal understanding has directly or indirectly hampered the scientific study of attention, as evident from \citet{chun2011taxonomy}. The concept of “Attention” has been studied under many lenses, from philosophy, psychology, cognitive sciences, neuroscience to computer science. The core ideology behind attention can be briefly stated as adaptive control of limited “computational” resources \citep{lindsay2020attention}, but it could be further generalized to any resources and not just computational. Although the idea behind the concept is simple, the attempts to rigorously define and unify it from different angles have proven difficult and thus there has been no global consensus. Therefore the view that attention is a unitary mechanism has been abandoned and is considered as a characteristic property of multiple constructs and mechanisms \citep{chun2011taxonomy}. This has been true under psychology, neuroscience, and brain sciences in general, where attention has been studied as one of the important pillars of their fields. 

Although the brain has been viewed as an information processing center where different parts “compute” different aspects of the needs of the living being, attention being one of it, the focus on actually understanding the algorithm behind has not been very clear. Computer scientists started paying attention to “attention” very recently and the motivation behind was to improve the performance of artificial neural networks \citep{DBLP:journals/corr/BahdanauCB14}. As of now the relationship between the study of attention from the point of view of brain sciences and its use as a tool to enhance artificial neural networks is unclear at present. Part of this paper is an attempt to bring light to this relation and justify whether studying attention from the brain science angle would be fruitful for the computational view and vice versa. 

\subsection{Origin of Attention in Brain}
The brain can be broadly divided into the cerebrum, the cerebellum, and the brainstem. The neural circuits underlying attention are primarily believed to be in the brain stem. These neuronal networks dynamically control the information flow into the thalamus and later onto the cortex \citep{lindsay2020attention}. This is known as the attentional bottleneck \citep{tombu2011unified}. Although one can localize the sources of attention in the brainstem networks, its effects are widespread and consist of different feedback loops. Attention can impact any part of the brain, not limited to the primary sensory areas and circuits governing emotions. Therefore, one cannot study it as a function of a localized area of the brain. 

We can interpret the attentional networks in terms of sustaining the alert state, directing towards salient sensory information, and executive control, which resolves the conflict among the competitive parts of the brain that might be active at the same time \citep{raz2004anatomy}. The directing network relies heavily on the parietal systems (part of the cerebral cortex, which primarily processes extrasensory information like touch, temperature), including the temporal-parietal junction and the superior parietal lobe. It is involved in both orientations of visual information and stimuli in other modalities (visual cortex and auditory cortex). The alerting network depends on thalamic areas (part of the forebrain which relays all sensory and motor signals), locus coeruleus, and other subcortical areas. The executive part of the attention network depends on the anterior cingulate and lateral areas of the prefrontal cortex (the part which handles decision-making and executive control of the brain) \citep{raz2004anatomy}.

Additionally, diffused neuromodulatory systems play an essential role in the control of general attention mechanisms. It consists of chemical networks of various neurotransmitters released by particular neural circuits. For example, neuroscientists suspect that norepinephrine, acetylcholine, and dopamine modulate alertness, orienting to salient information, and executive control of attention, respectively \citep{posner2008measuring}.

\subsection{Types of Attention}

\begin{figure}[h]
    \centering
    \includegraphics[width=0.619\textwidth]{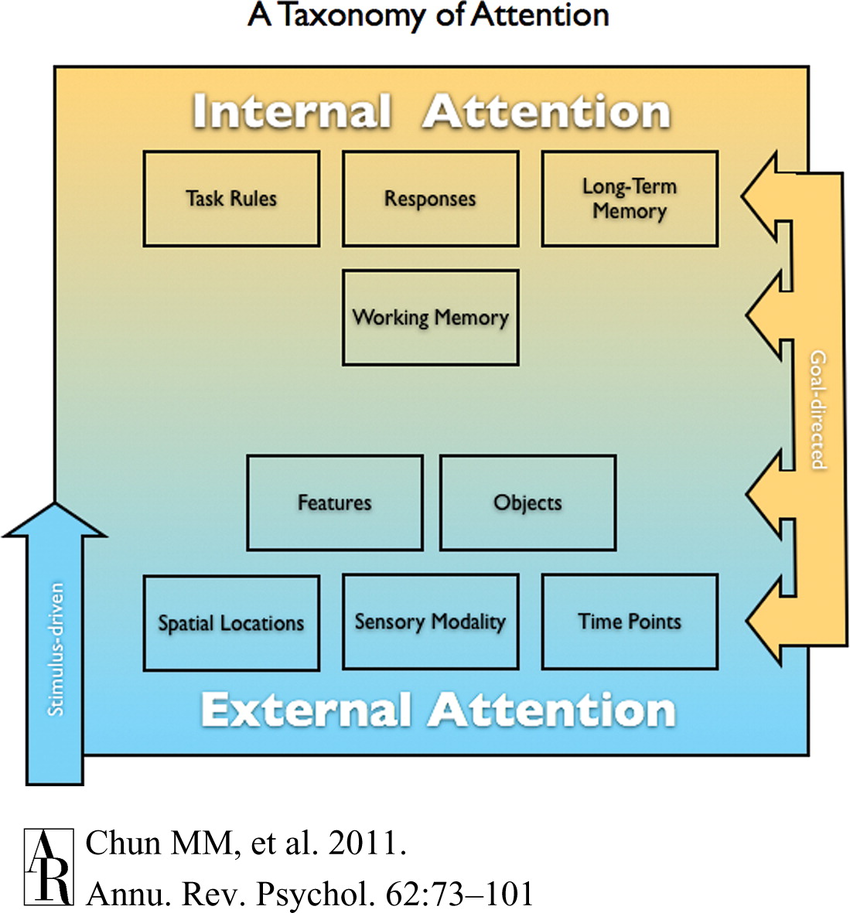}
    \caption{A schematic overview of external and internal attention, image taken from \citet{chun2011taxonomy}}
    \label{fig:attentaxon}
\end{figure}

\subsubsection{External Vs. Internal}
Attention can be broadly studied in two halves, although they are inseparable. External attention mainly refers to the attention applied directly to the sensory streams. It can be applied to raw sensory information instantaneously or after preprocessing in lower layers of the nervous system. It develops topological maps for localizing important spatial locations, temporal significance, objects, and modality-specific features directly from the incoming sensory signals in a bottom-up fashion. Therefore it’s also known as “Bottom-Up” attention. These maps help in generating saliency for important aspects of input signals thus allowing to direct the limited resources of the brain in processing only the salient information \citep{chun2011taxonomy}. 

Internal attention, unlike the former, is difficult to define in a unified fashion. But it mainly refers to how the executive functions of the brain command, control, and direct the internal resources of the cortices. This includes but is not limited to encoding and decoding of long term memories, managing working memory, task-relevant, and task-irrelevant rules and processing. The higher layers directly help in modulating the information entering from the lower hierarchy of neural information processing levels. Thus it’s also known as “Top-Down” attention as it controls the flow of information entering the brain in a top-down fashion \citep{chun2011taxonomy}. A schematic overview is presented in Figure \ref{fig:attentaxon}.

\subsubsection{Spatial Vs. Feature}

Spatial and Feature Attention can be considered as part of external attention mechanisms. In a broad sense, Spatial Attention determines how to give importance to spatial locations in the environment \citep{chun2011taxonomy}. This is a key example of the attentional bottleneck problem. It solves this by using the bottom-up saliency maps obtained from lower sensory layers. Spatial Attention is a core part of the vision system of the brain, where it helps in deploying fovea through saccadic movements. But principles of Spatial Attention are also applicable to other modalities in general.

Unlike Spatial Attention, which is about focusing on a particular location of the sensory space, Feature Attention is global \citep{lindsay2020attention}. It helps in determining orientation to certain features or objects that can be captured beyond modality, spatial, and temporal aspects. Features can be considered as points in dimensions specific to modalities, such as color, temperature, and pitch. Feature Attention controls and improves the signal processing of feature selective circuits of the cortical regions \citep{chun2011taxonomy}.


\section{Towards Unifying Attention in AI and Neuroscience}
\label{sec:UnifyingAINS}
Deep neural networks (DNNs) are very different from biological neural networks (BNNs). The electrochemical complexity of BNNs from ion channels to neurotransmitters has been completely abstracted by DNNs with simple linear and non-linear transformations as evident from \citet{marblestone2016toward}. Perhaps the only common theme between both of them is the connectionist ideology. But the motivation behind them has remained the same, to capture the learning and information processing abilities of BNNs in a simplistic manner. In general, it’s to reverse engineer the functional aspects of the computational capabilities of the brain while remaining indifferent to the structural aspects. Although the functioning of the brain is still a black box (which can be mirrored with the interpretability issues with deep learning algorithms), neuroscientists have helped us unlock many mysteries of the brain. Some of them have directly corresponded with the functioning of DNN architectures. For example, the activations of the deep convolutional neural network have been determined to be correlated with the neural activity of the visual cortex \citep{yamins2016using}. In fact, CNNs can be studied as a functional model of the vision system of the brain as argued by \citet{lindsay2020convolutional}.

The dialogues between varied disciplines such as computer science, neuroscience, and cognitive sciences however lack an interdisciplinary construct for understanding and unifying the common ideas in these fields. We posit that a conceptual framework for understanding these ideas on a unified platform is necessary.
Early researchers faced a similar challenge during the rise of their respective fields. 
And their pioneering solutions can still be used for analyzing modern issues concerning deep learning and neuroscience as shown in \citet{hamrick2020levels}. 

Conceptual frameworks are mental models that align researchers’ understanding of how their work and philosophies fit into the larger goals of their field and science more broadly \citep{hamrick2020levels}. One of the most influential of such models is Marr’s Levels of analysis, introduced by Marr in 1982 which pioneered the research in cognitive neuroscience. It consists of a three-layer hierarchy for analyzing computational aspects of the brain’s functionality:  

\begin{enumerate}
    \item \textbf{Computational level}: It describes the goal of a system, its I/O relationship, and possibly if it can be formulated mathematically. 
    \item \textbf{Representational level}: It refers to possible representations, algorithms, and data structures to be used to solve the computational goal. 
    \item \textbf{Implementation level}: The lowest level is about the actual implementation of the solution which serves the goal. It could be either physically embedded in hardware, or simulated in software. At this level, we can ask questions like how a neural circuit in our brain might represent the data structures or programs using spike codes.
\end{enumerate}

We can see that Marr's levels are clearly influenced by software and hardware abstraction layers in computer science as evident from \citet{love2019levels}. Although Marr's model is not as detailed as mental frameworks used by computer scientists, it's usefulness lies in the flexibility neuroscientists need. But one thing is clear that as we go deeper into the levels, more information is introduced regarding bare computational or information processing requirements whereas as upper levels discusses abstract concepts which could be manifested in multiple ways for example in vivo or in silico. One way to analyse attention in general using Marr's model could be:

\begin{enumerate}
    \item \textbf{Computational Level} :
    \begin{itemize}
        \item To efficiently determine the most salient of the input signals.
        \item To dedicate the limited computational resource in most efficient way via the attentional bottleneck.
    \end{itemize}
    \item \textbf{Representation Level}:
    \begin{itemize}
        \item Using Differentiable Programming models. \citep{innes2019differentiable}.
        \item Using object oriented deep learning \citep{liao2017object}.
        \item With Memory Augmented Networks \citep{graves2016hybrid}.
        \item With information rich representations like Probabilistic Graphical Models \citep{koller2009probabilistic}.
    \end{itemize}
    \item \textbf{Implementation Level}:
    \begin{itemize}
        \item Brain:
        \begin{itemize}
            \item Spike codes and Population codes mediated by Biological Neural Circuits \citep{brendel2020learning}.
            \item Reinforcement Learning and Diffused Modulatory Systems in brain \citep{avery2017neuromodulatory}.
        \end{itemize}
        \item Computer:
        \begin{itemize}
            \item To efficiently implement and scale the attention operation in hardware. E.g. On a CPU, a single GPU, multiple GPUs or TPUs \citep{abadi2016tensorflow}.
            \item Neuromorphic computing with spiking neural networks (SNNs) and Spiking Neural Processing Units (NPUs) \citep{schuman2017survey}.
        \end{itemize}
    \end{itemize}
\end{enumerate}

At the computational level, we formulate the attention mechanism's goal in general, which is to determine salient signals via the attentional bottleneck efficiently. To compute attention, process the input signals so that none of the information rich signals are lost. At the same time, contextually exclude irrelevant information. Here context recognition will be detrimental for attention as the saliency of a particular signal directly depends on its context. At the representation level, we back the differentiable programming models \citep{innes2019differentiable} as gradient-based learning is state of the art and highly efficient approach for solving any learning algorithm. However, it is worthwhile exploring other promising and non-trivial approaches for computing attention like object-oriented deep learning \citep{liao2017object}. For deploying attention, the algorithm must be able to separate the required signal from the irrelevant ones. We believe disentangling this information from the input signal representations \citep{higgins2018towards} will significantly improve this task. At the representation level, obtaining rich representations of the input signal is also a bottleneck of the current artificial neural networks. Object-oriented deep learning is one of the recent methods for obtaining disentangled representations, and hence we suggest the same. At the implementation level, we describe how brains and computers can compute the same functionality, albeit having a completely different underlying basis of their respective mechanisms.

Bridging the gap between varied disciplines by placing the ideas on common conceptual grounds will be the key to solve the problems which none of the disciplines can solve individually. Marr’s levels of analysis might not be the most suitable or the only such model for grounding interdisciplinary problems arising in computer sciences and neurosciences. However, such conceptual frameworks help us in comparing various learning algorithms with constraints arising in neuroscience and also formalize the areas of disagreement. This could be the way forward towards the systematic unification and analysis of ideas arising from multiple fields.




\section{Conclusion}

Attention is a broad concept that is being studied across multiple fields not just limited to AI and Neuroscience. Although many of the ideas from these different fields do not necessarily intersect with each other, there does exist a core commonality of adaptive control of limited resources and the attentional bottleneck. Instead of having seemingly disconnected dialogues between varied disciplines we suggest grounding the ideas on common conceptual frameworks like Marr’s levels of analysis or related mental models. Having a common basis won’t only bring clarity to solving problems requiring multidisciplinary views but also bring out possible errors in our thought processes while modeling such problems.

\section{Acknowledgements}
We would like to thank Dr. Anirban Dasgupta, Mr. Rachit Chhaya and the anonymous reviewers for their feedback on this paper. 

\bibliography{neurips_2019}
\bibliographystyle{acl_natbib}
\newpage

\end{document}